\documentclass[10pt,twocolumn,letterpaper]{article}

\usepackage{cvpr}
\usepackage{times}
\usepackage{epsfig}
\usepackage{graphicx}
\usepackage{amsmath}
\usepackage{amssymb}
\usepackage{multirow}
\usepackage{booktabs}
\usepackage{threeparttable}

\usepackage[pagebackref=true,breaklinks=true,letterpaper=true,colorlinks,bookmarks=false]{hyperref}

\cvprfinalcopy % *** Uncomment this line for the final submission

\def\cvprPaperID{7229} % *** Enter the CVPR Paper ID here

\newcommand{\fig}{Figure }
\newcommand{\tab}{Table }
\newcommand{\eqn}{Eq. }

\ifcvprfinal\fi
\begin{document}

%%%%%%%%% TITLE
\title{Solving Raven's Progressive Matrices with Neural Networks}

\author{Tao Zhuo\\
National University of Singapore\\
Singapore\\
{\tt\small zhuotao@nus.edu.sg}
\and
Mohan Kankanhalli\\
National University of Singapore\\
Singapore\\
{\tt\small mohan@comp.nus.edu.sg}
}

\maketitle

\begin{abstract}

Raven's Progressive Matrices (RPM) have been widely used for Intelligence Quotient (IQ) test of humans. In this paper, we aim to solve RPM with neural networks in both supervised and unsupervised manners. First, we investigate strategies to reduce over-fitting in supervised learning. We suggest the use of a neural network with deep layers and pre-training on  large-scale datasets to improve model generalization. Experiments on the RAVEN dataset show that the overall accuracy of our supervised approach surpasses human-level performance. Second, as an intelligent agent requires to automatically learn new skills to solve new problems, we propose the first unsupervised method, Multi-label Classification with Pseudo Target (MCPT), for RPM problems. Based on the design of the pseudo target, 
MCPT converts the unsupervised learning problem to a supervised task. Experiments show that MCPT doubles the testing accuracy of random guessing \eg 28.50\% vs. 12.5\%. Finally, we discuss the problem of solving RPM with  unsupervised and explainable strategies in the future.

\end{abstract}

\section{Introduction}

\begin{figure}[!t]
	\centering
	\includegraphics[width=0.45\textwidth]{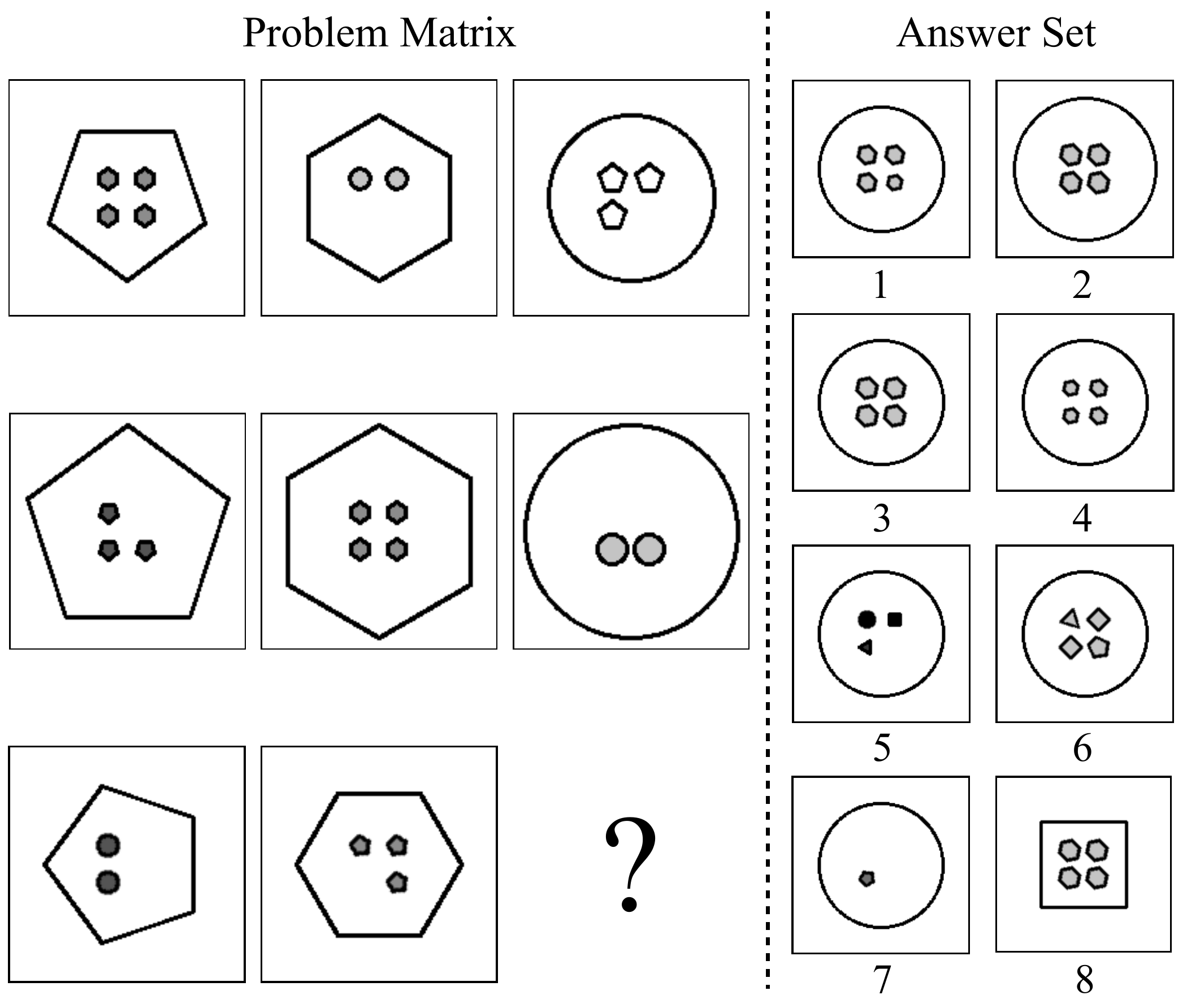}
	\caption{An example of the RPM problem~\cite{CVPR2019_Zhang}. The test taker has to select the best choice in the answer set to complete the problem matrix. Some shared rules by the first and second rows include: (1) Shape: pentagon, hexagon, and circle. (2) Number of inside objects: 2, 3, and 4. (3) Size: inside objects are of the same size. Therefore, the missing piece in the third row should be a circle shape that contains 4 hexagons of the same size. The candidate answers can be 2 or 3. By further considering the size of the circle, the best answer is 3.}
	\label{fig_rpm}
\end{figure}

Raven's Progress Matrices (RPM) problem~\cite{Book2006_Domino, ECPA1938_Raven,IJCAI2015_Wang,ICML2018_Santoro,CVPR2019_Zhang,NeurIPS2019_Zhang} is one of the most popular instruments to measure abstract reasoning and fluid intelligence through non-verbal questions. As in the example shown in \fig \ref{fig_rpm}, the problem matrix in an RPM problem consists of several visual geometric designs with a missing piece. Given eight candidate choices as the answer set, the test taker has to figure out the logical rules hidden in the problem matrix~\cite{CVPR2019_Zhang}, and selects the best choice from the answer set that satisfies these hidden rules~\cite{Book2006_Domino}. 

Unlike other computer vision tasks, such as image classification~\cite{ICCV2015_He}, video action recognition~\cite{CVPR2017_Carreira,CVPR2018_Hara,MM2019_Zhuo} and Visual Question Answering (VQA)~\cite{ICCV2015_Antol,CVPR2019_Zellers}, RPM lies directly at the heart of human intelligence~\cite{CVPR2019_Zhang}. The RPM problem is independent of language, reading and writing skills, as well as of cultural backgrounds. It provides a simple yet effective way to evaluate the test taker's reasoning ability~\cite{Psy1990_Carpenter, Cogn2000_Raven}. Thus, RPM has been widely used to test human cognitive abilities and Intelligence Quotient (IQ). 

In the field of Artificial Intelligence (AI), one of the most important goals is to make machines with reasoning ability similar to that of humans, and hence RPM has received increasing attention in recent years. It is expected that the RPM problem can help understand and improve the reasoning ability of current computer vision systems. To solve RPM with machines, Wang and Su~\cite{IJCAI2015_Wang} introduced an abstract representation method for RPM by using the first-order logic formulae, and they applied three categories of relations (\ie unary, binary, and ternary) for automatic RPM generation. Santoro \etal~\cite{ICML2018_Santoro} developed a Procedurally Generated Matrices (PGM) dataset by instantiating each rule with a relation-object-attribute. To improve the reasoning ability, they proposed a Wild Relation Network (WReN) for inferring inter-panel relationships between the problem matrix and the answer set.

To further push the limits of modern visual systems' reasoning ability, Zhang \etal~\cite{CVPR2019_Zhang} generated a more complicated RPM dataset, which is called Relational and Analogical Visual rEasoNing (RAVEN) in homage to John Raven for his pioneering work~\cite{ECPA1938_Raven}. Besides, they provided benchmark results of human performance to measure the reasoning ability of machines. Moreover, a module called Dynamic Residual Tree (DRT)~\cite{CVPR2019_Zhang} is proposed to leverage the structure annotations for each RPM problem. Recently, a new permutation-invariant model CoPINet~\cite{NeurIPS2019_Zhang} achieves the state-of-the-art performance, by borrowing the idea of ``contrast effects'' from the fields of psychology, cognition, and education. 

In this paper, we attempt to solve RPM with neural networks employing both supervised and unsupervised strategies. First, we investigate the supervised learning strategy that considers the RPM problem as a multi-class classification problem. Different from the previous works~\cite{arxiv2017_Hoshen,ICML2018_Santoro,CVPR2019_Zhang,NeurIPS2019_Zhang} that focus on designing an effective neural network, we mainly analyze how to reduce over-fitting in RPM, which is an important commonly occurring problem in supervised learning. An interesting observation is: ImageNet~\cite{CVPR2009_Deng} pre-training is very effective in improving the model generalization, although images in RPM problems are quite different from the real-world images. In addition, a neural network with deep layers can also help reduce over-fitting and help increase the testing accuracy. It is worth mentioning that our supervised method with ResNet-50 backbone attains a testing accuracy of 86.26\%, which surpasses human-level performance (84.41\%) on the RAVEN dataset~\cite{CVPR2019_Zhang}. 

Second, as humans can solve RPM without any prior experience and supervision, a natural question arises: \emph{whether machines have reasoning ability comparable to that of humans who can solve RPM without any labeled data?} In order to attempt to answer this question, we propose a novel method named Multi-label Classification with Pseudo Target (MCPT) to solve RPM in an unsupervised manner, which makes the neural network more human-like. Given an RPM problem, one observation is that the third row with the correct answer has to satisfy the same rules shared between the first and second rows in each problem matrix, as illustrated in \fig \ref{fig_rpm}. Since each RPM is built employing its own logical rules that are unknown, if we cannot extract rule-based features on each row in advance, it is not possible to find the answer by a Nearest Neighbor Search (NNS) strategy directly. Therefore, we first fill each candidate choice in the answer set to complete the missing piece, generating 10 rows in total. Then, we design a pseudo target of a two-hot vector with 10 dimension, in which only the first two elements are set to 1, while the rest are set to 0, corresponding to 10 rows. Based on such a pseudo target design, it is expected that the row with correct answer could be classified to the same label as that of the first two rows. Essentially, the RPM problem with unsupervised learning setting is converted to a supervised multi-label classification task with pseudo target, which is at the core of our MCPT method. Compared to the random guess strategy (12.5\%, as there are 8 candidate choices), our MCPT doubles the testing accuracy (28.50\%).

Our main contributions can be summarized as follows: 
(1) We suggest the use of a neural network with deep layers and pre-training on large datasets to improve model generalization in supervised learning. Experiments on the RAVEN dataset show that our approach achieves an average testing accuracy of 86.26\%, which surpasses human-level performance (84.41\%).
(2) We propose the first unsupervised deep learning method MCPT for RPM, which uses a pseudo target of a two-hot vector to convert the unsupervised learning to a supervised one. Extensive experiments show that MCPT effectively increases the testing accuracy by a large margin of 22\%, and it doubles the testing accuracy of random guess, \ie 28.50\% vs 12.50\%. 
In addition, we discuss machine intelligence and advocate more works on unsupervised and explainable learning strategies. {\bf Our code will be released.}

\section{Related Work}

\subsection{Different Forms of RPM}
The RPM problem was first published by John Raven for research purposes in 1938~\cite{ECPA1938_Raven}. There are currently three different forms of RPM for test takers: (1) Standard Progressive Matrices (SPM), which is the original form. All items in SPM are presented in black ink and white background. (2) Colored Progressive Matrices (CPM), which is designed on a colored background. CPM is simpler than SPM, and thus it is often used for children aged 5 through 11 years of age, the elderly, the mentally, and physically impaired individuals. (3) Advanced Progressive Matrices (APM), which consists of more items than SPM. APM is more difficult than SPM and it is designed for adults. In this work, we use the RAVEN dataset~\cite{CVPR2019_Zhang} for performance analysis, which follows the rules~\cite{Psy1990_Carpenter} of APM.

\subsection{Different Models for RPM}
It is widely believed that RPM problems are highly correlated with general intelligence. Many computational models have been developed to study the reasoning ability of machines by solving RPM problems. To figure out how different representations can contribute to SPM problems, Kunda \etal~\cite{AAAI2009_Kunda} used propositional, imagistic, and multimodal representations for solving RPM with simple classifiers. Later, various methods~\cite{AAAI2014_Mcgreggor,AI2014_Mcgreggor,IJCAI2018_Mekik} were developed to solve RPM by computing the feature similarity of images. Besides, to explore structural relationships in RPM problems, Shegheva and Goel~\cite{AAAI2018_Shegheva} proposed a structural affinity method that used graphical models for model learning and pattern recognition. 

Recently, with the success of deep learning in various computer vision tasks, such as object detection~\cite{ICCV2015_He}, segmentation~\cite{ECCV2018_Chen} and VQA~\cite{ICCV2015_Antol,CVPR2019_Zellers}, different deep learning models have been designed to solve RPM in the past  decade. Hoshen and Werman~\cite{arxiv2017_Hoshen} first trained a CNN model to measure the IQ of neural networks in a simplistic evaluation environment. Santoro ~\etal proposed a WReN~\cite{ICML2018_Santoro} network that formulates pair-wise relations between the problem matrix and each individual choice in an embedding space, independent of the other choices. They further studied the generalization of their model. More recently, a new approach called Dynamic Residual Tree (DRT)~\cite{CVPR2019_Zhang} is proposed to leverage the structure annotations for each RPM problem. In addition, a new method CoPINet~\cite{NeurIPS2019_Zhang} borrows the idea of ``contrast effects'' from the fields of psychology, cognition, and education. By combining contrasting, perceptual inference, and permutation invariance, CoPINet achieves the state-of-the-art performance. 

\section{Our Approaches}
Let $\mathbf{x}$ be a problem matrix that consists of $m$ images, $\mathbf{y}$ be the corresponding answer set that consists of $n$ candidate choices. Solving an RPM problem instance is to select the best answer $\mathbf{y}_z$ to complete the problem matrix $\mathbf{x}$, where $z$ is the index of $\mathbf{y}_z$ and $z \in \{1, 2, \cdots, n\}$. Next, we introduce our supervised and unsupervised methods for this problem.

\subsection{Supervised Learning Approach}
Unlike previous supervised works that focus on designing an effective neural network, we mainly investigate how to reduce over-fitting in RPM, as it is a commonly occurring problem. Before discussing the effect of model pre-training and deep layers, we first introduce the problem formulation.

\subsubsection{Problem Formulation}
Given $T$ training samples $\{(\mathbf{x}_i, \mathbf{y}_i, z_i)\}_{i=1}^{T}$ with $\mathbf{x}_i \in X$, $\mathbf{y}_i \in Y$ and $z_i \in Z$, we aim to learn a projection function $f$ over $(X, Y)$ and $Z$. In our implementation, all images in the problem matrix and its corresponding answer set are stacked together, and then fed into a neural network for the final answer prediction. The supervised learning method can be formulated as: 
\begin{equation}
Z = f(\phi(X \cup Y); \mathbf{w}),
\label{eq_sup}
\end{equation}
where $\cup$ denotes the concatenation operation that stacks all images in the problem matrix and the answer set; $\phi$ represents the image features over $(X \cup Y)$; $\mathbf{w}$ is the hyper parameter set to learn. The RPM problem can then be considered as a multi-class classification task.

During the model training stage, we compute the cross-entropy loss $L_{ce}(\mathbf{w})$ for each RPM problem as: 
\begin{equation}
L_{ce}(\mathbf{w}) = -\sum_{c=1}^{n} z_c \log(p_c),
\end{equation}
where $p_c=f_c(\phi(\mathbf{x}_i \cup \mathbf{y}_i); \mathbf{w})$ is the model's estimated probability for the class with label $c$. During the testing stage, the choice with the highest output probability is considered to be the final answer.

\subsubsection{Model Pre-training and Deep Layers}
Since ResNet~\cite{CVPR2016_He} has demonstrated  good performance and shown its surprising effectiveness in pattern recognition, we use it to study the reasoning ability of machines. We replace the input dimension (for all 16 images in an RPM question) of the first convolutional layer and output dimension (for 8 candidate choices) of the fully connected layer. Besides, a dropout layer is added before the fully connected layer to reduce over-fitting. For supervised learning approaches, model initialization and the number of layers are very important. In the next, we study the effectiveness of ImageNet pre-training~\cite{CVPR2009_Deng} and deep layers in solving RPM problems.

ImageNet pre-training has been widely used in many computer vision tasks for model initialization~\cite{ICCV2019_He}, \eg, semantic segmentation~\cite{ECCV2018_Chen} and video action recognition~\cite{CVPR2017_Carreira,CVPR2018_Hara}, as it is an effective strategy to reduce over-fitting and speed up model convergence. However, unlike the real-world images in the ImageNet dataset, the images in RPM consist of geometric shapes (see \fig \ref{fig_rpm}). So, RPM images are quite different from ImageNet images. {\bf Is ImageNet pre-training still helpful to RPM problems?}

In addition, as the results reported in the previous work \cite{CVPR2019_Zhang}, DRT achieves the best performance with ResNet-18 backbone after testing several ResNet variants without model pre-training. However, in many computer vision tasks~\cite{CVPR2017_Carreira,ECCV2018_Chen,CVPR2018_Hara}, neural networks with deep layers often show superior performance. {\bf Can we further improve the performance with deep layers in RPM?}

\begin{figure}[!t]
	\centering
	\includegraphics[width=0.4\textwidth]{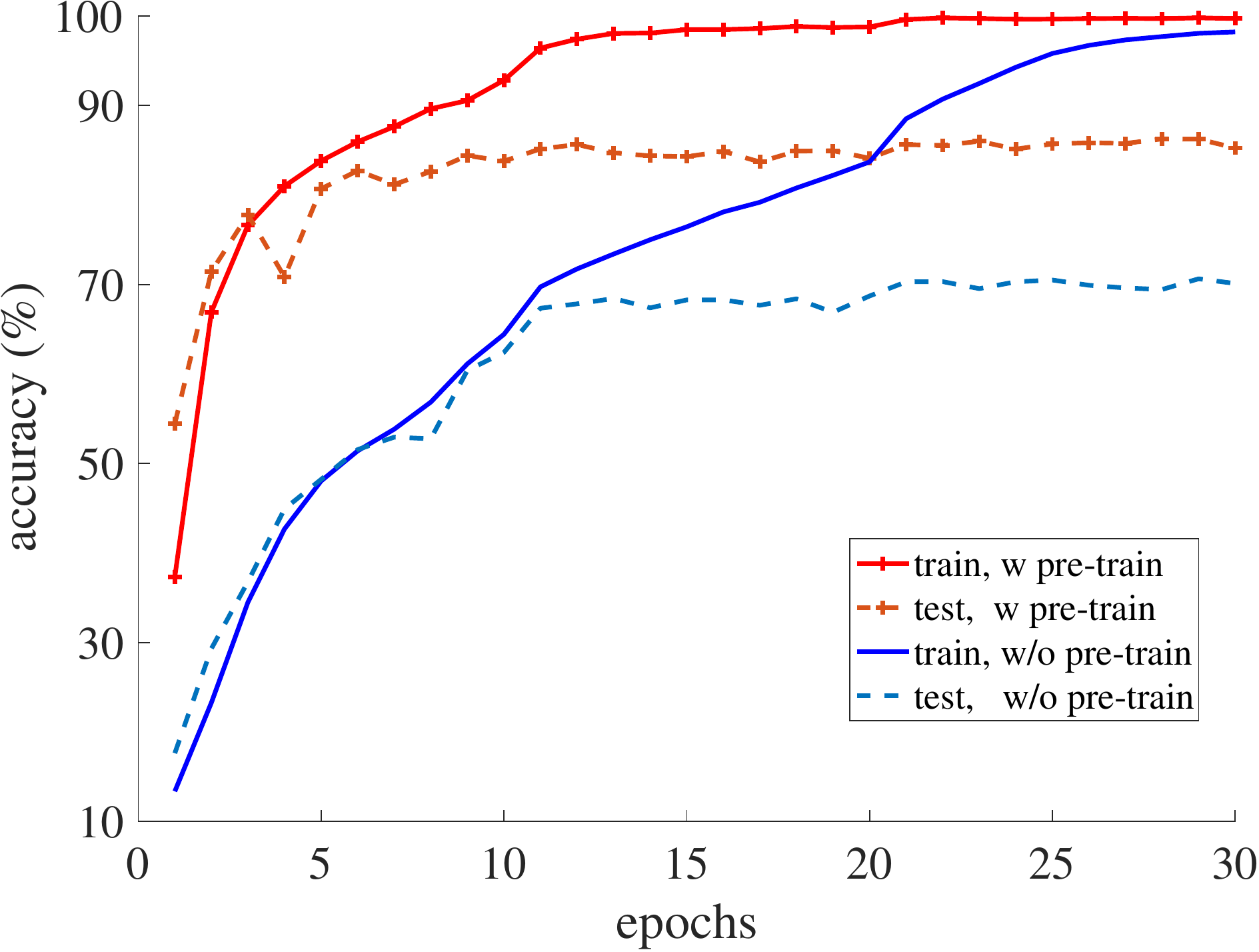}
	\caption{The effect of ImageNet pre-training with a ResNet-50 backbone in our experiments.}
	\label{fig_pretrain}
\end{figure}

To answer the above questions, we conduct experiments on 4 representative ResNet variants using the RAVEN dataset~\cite{CVPR2019_Zhang}: ResNet-18, ResNet-34, ResNet-50, and ResNet-101. \tab \ref{tbl_diffres} shows the testing accuracy of each model. We use \fig \ref{fig_pretrain} to further illustrate the effect of ImageNet pre-training with the ResNet-50 backbone. Our observations can be summarized as follows:
\begin{itemize}
\item ImageNet pre-training can help reduce over-fitting and speed up model convergence in RPM, although their image types are quite different. From \tab \ref{tbl_diffres}, we can observe that the testing accuracy of neural networks can be substantially improved with ImageNet pre-training, \eg, up to 33.79\% for ResNet-101. ResNet-18 consists of a few layers only, its performance does not change much with or without model pre-training. In contrast, the accuracy increase in other three deeper models is notable. \fig \ref{fig_pretrain} shows that ImageNet pre-training speeds up the model convergence (with pre-training: 12 epochs, without pre-training: 30 epochs) and improves the model generalization (\ie 86.26\% vs 70.63\%, about 16\% increase). 	
	
\item Neural network with deep layers is useful to improve the model generalization, but it will be saturated as the depth increases. Without ImageNet pre-training, ResNet-18 achieves the best performance, which means neural network with deeper layers does not improve the testing accuracy by using randomly initialized weights. In contrast, based on the ImageNet pre-training, ResNet-50 improves the testing accuracy by about 10\%, when compared to ResNet-18 without pre-training. However, when the neural network becomes deeper, the performance gets saturated, as ResNet-101 does not obtain further improvement. 	
\end{itemize}

Based on the above observations, we suggest the use of a neural network with deep layers and pre-training model on large-scale datasets to reduce over-fitting and further improve the testing accuracy for RPM problems.

\begin{table}[tb]
\centering
\caption{Testing accuracy of each model without (w/o) or with (w) ImageNet pre-training. R denotes the ResNet backbone.}
\begin{tabular}{lcccc} \toprule 
Model               & R-18      & R-34      & R-50        & R-101     \\ \midrule  
w/o  pre-training   & 77.18     & 54.34     & 70.63       & 51.96     \\ 
w pre-training      & 76.38     & 75.87     & \bf{86.26}  & 85.75     \\ \bottomrule
\end{tabular}
\label{tbl_diffres}	
\end{table}

\begin{figure*}[!t]
\centering
\includegraphics[width=0.7\textwidth]{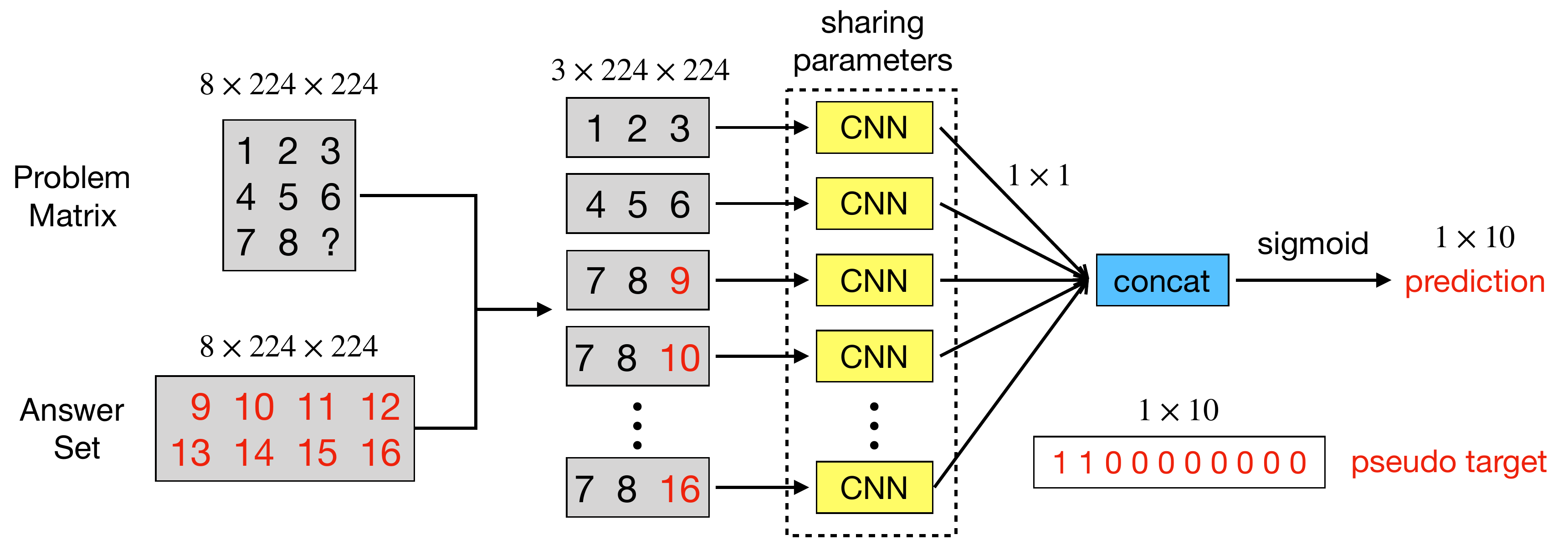}
\caption{An overview of MCPT. The numbers in the boxes of the problem matrix and the answer set represent the corresponding images in an RPM problem. During the training stage, we use a two-hot vector with 10 dimensions as the pseudo target. During the testing stage, we remove the first two elements of the prediction and estimate the final answer within the output probabilities of the rest 8 values.}
\label{fig_unsup}
\vspace{-1em}
\end{figure*}
\subsection{Unsupervised Learning Approach}
\subsubsection{Motivation}
For an intelligent agent, automatically learning new skills is very important, as it can help solve new problems in new environments. Humans are able to solve RPM problems without any experience and supervision. If machines could have reasoning ability comparable to that of humans, an interesting question arises: can machines automatically solve RPM without any labeled data? To try to answer this question, we attempt to solve RPM in an unsupervised manner. 

\subsubsection{Technical Challenges}
Given $T$ samples $\{(\mathbf{x}_i, \mathbf{y}_i)\}_{i=1}^{T}$ with $\mathbf{x}_i \in X$ and $\mathbf{y}_i \in Y$, we aim to design a model that can automatically find correct answers for $(X, Y)$. As the example shown in \fig \ref{fig_rpm}, one observation is that the three rows (with the correct answer) in the problem matrix have to satisfy the same rules. If there exists a rule-based feature extraction model on each row in advance, we can find the correct answer directly by a NNS (Nearest Neighbor Search) approach. However, the logical rules in different RPM problems are different and unknown. What's worse is - the number of rules in different RPM problems is also different and unknown as well. To the best of our knowledge, there is no rule-based feature extraction method available for RPM problems. Therefore, the NNS strategy cannot be employed to solve RPM.

Recently, to automatically cluster similar entities (\eg, images in ImageNet) into the same groups, some unsupervised deep clustering methods~\cite{ECCV2018_Caron,NeurIPS2017_Guo,ICML2016_Xie,CVPR2019_Yang} propose to learn feature representations and group assignments simultaneously using deep neural networks. For example, Caron \etal~\cite{ECCV2018_Caron} iteratively used K-means to find the shared centroids and produced pseudo-labels on different images, and then used these pseudo-labels to learn the network parameters. However, RPM problems are different from those traditional clustering tasks. Each RPM problem $(\mathbf{x}_i, \mathbf{y}_i)$ is built employing its own logical rules, which means there are no shared centroids among different RPM problems. As a result, the unsupervised deep clustering methods cannot be used to solve RPM. 

\subsubsection{The MCPT Approach}
We now introduce the Multi-label Classification with Pseudo Target (MCPT) approach that solves RPM in an unsupervised manner. Without any labeled data for supervision, the target to be learned is unknown in unsupervised learning. To tackle this issue, we observe that the three rows in the problem matrix have to satisfy the same rules. Based on such an observation, we design a two-hot vector as the pseudo target, which converts the unsupervised learning problem to a supervised one. 

As illustrated in \fig \ref{fig_unsup}, by independently filling each candidate answer to the missing piece, 10 image rows can be generated. One prior is that the correct answer $(\mathbf{x}_{i, 7}, \mathbf{x}_{i, 8}, \mathbf{y}_{i, j})$, $j \in \{1, 2, \ldots, 8\}$, has to satisfy the same rules shared by the first row  $(\mathbf{x}_{i, 1}, \mathbf{x}_{i, 2}, \mathbf{x}_{i, 3})$ and the second row $(\mathbf{x}_{i, 4}, \mathbf{x}_{i, 5}, \mathbf{x}_{i, 6})$ in an RPM problem. Then, we design a pseudo target of a two-hot vector, in which the first 2 elements are set to 1, while the rest 8 are set to 0, corresponding to 10 rows. Based on such a pseudo target design, it is expected that the row with correct answer could be classified to the same label as that of the first two rows. 

We use a CNN network to extract the features of each row, followed by a dropout layer and a fully connected layer (output dimension is 1) to predict the ranking score for each row. By concatenating those 10 ranking scores together, we use a sigmoid function for normalization. Then, the proposed unsupervised learning method can be written as:
\begin{equation}
\tilde{Z} = g(\varphi(X \vee Y); \mathbf{w}),
\label{eq_unsup}
\end{equation}
where $\tilde{Z}$ is the pseudo target for all RPM problems; $g$ is a projection function; $\varphi$ is the image features; $\{\mathbf{r}_k\}_{k=1}^{10}$ denotes reorganized image rows of $(X \vee Y)$, \ie $\mathbf{r}_1=(\mathbf{x}_{i, 1}, \mathbf{x}_{i, 2}, \mathbf{x}_{i, 3})$, $\mathbf{r}_2=(\mathbf{x}_{i, 4}, \mathbf{x}_{i, 5}, \mathbf{x}_{i, 6})$, $\mathbf{r}_{j+2}=(\mathbf{x}_{i, 7}, \mathbf{x}_{i, 8}, \mathbf{y}_{i, j})$, where $j \in \{1, 2, \cdots, 8\}$. Compared to the true target $Z$ in \eqn \ref{eq_sup}, $\tilde{Z}$ is only a prior, which converts  unsupervised learning to a supervised problem, \ie Multi-label Classification with Pseudo Target (MCPT). 

For model training, we adopt the widely used loss function Binary Cross Entropy (BCE) $L_{bce}(\mathbf{w})$ in multi-label classification, which is formulated for each RPM as:
\begin{equation}
L_{bce}(\mathbf{w}) = - \sum_{k=1}^{10} [\tilde{z} \log(p_k) + (1-\tilde{z})\log(1-p_k)],
\end{equation}
where $\tilde{z} \in \{0, 1\}$ is the pseudo label for each row $\mathbf{r}_k$ in \fig \ref{fig_unsup}; $p_k=g(\varphi(\mathbf{r}_k); \mathbf{w})$ is the model's estimated probability for each row $\mathbf{r}_k$. During the testing stage, we remove the first two elements of the prediction, then the maximum output of the rest 8 values (corresponding to 8 candidate choices) is the final answer.

\section{Experiments}
\label{sec_exp}
In this section, we study the performance of both our supervised and unsupervised approaches. Next, we introduce the RAVEN dataset and implementation details, and then we report the performance of our approaches. 

\begin{figure*}[!t]
	\centering
	\includegraphics[width=1\textwidth]{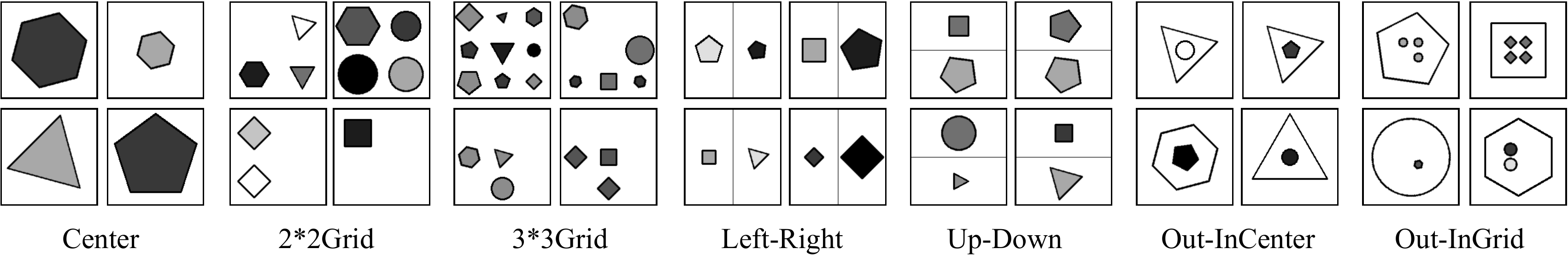}
	\caption{Examples of 7 different figure configurations in the RAVEN dataset.}
	\label{fig_type}
\end{figure*}

\subsection{Dataset}
To demonstrate the effectiveness of our methods, we evaluate them on the latest RAVEN dataset~\cite{CVPR2019_Zhang}. In total, the RAVEN dataset consists of 1,120,000 images and 70,000 RPM problems, equally distributed in 7 distinct figure configurations: \emph{Center}, \emph{2*2Grid}, \emph{3*3Grid}, \emph{Left-Right} (\emph{L-R}), \emph{Up-Down} (\emph{U-D}), \emph{Out-InCenter} (\emph{O-IC}) and \emph{Out-InGrid} (\emph{O-IG}), as shown in \fig \ref{fig_type}. In addition, there is an average of 6.29 rules for each problem. To evaluate the reasoning ability of machines, the RAVEN dataset also contains the results of human-level performance. More details of the RAVEN dataset are described in~\cite{CVPR2019_Zhang}.  

\begin{table*}[!t]
	\centering
	\caption{Testing accuracy of different models in supervised manner. \emph{Avg} denotes the average accuracy of each model.}
	\begin{tabular}{lcccccccc} \toprule
		Method             & \emph{Avg}   & \emph{Center} & \emph{2*2Grid}  & \emph{3*3Grid}
		& \emph{L-R}   & \emph{U-D}   & \emph{O-IC}  & \emph{O-IG}  \\  \midrule
		LSTM               & 13.07 & 13.19  & 14.13   & 13.69   & 12.84 & 12.35 & 12.15 & 12.99 \\
		WReN               & 14.69 & 13.09  & 28.62   & 28.27   & 7.49  & 6.34  & 8.38  & 10.56 \\
		CNN                & 36.97 & 33.58  & 30.30   & 33.53   & 39.43 & 41.26 & 43.20 & 37.54 \\
		ResNet-18+MLP      & 53.43 & 52.82  & 41.86   & 44.29   & 58.77 & 60.16 & 63.19 & 53.12 \\ \hline
		LSTM+DRT           & 13.96 & 14.29  & 15.08   & 14.09   & 13.79 & 13.24 & 13.99 & 13.29 \\
		WReN+DRT           & 15.02 & 15.38  & 23.26   & 29.51   & 6.99  & 8.43  & 8.93  & 12.35 \\
		CNN+DRT            & 39.42 & 37.30  & 30.06   & 34.57   & 45.49 & 45.54 & 45.93 & 37.54 \\
		ResNet-18+MLP+DRT  & 59.56 & 58.08  & 46.53   & 50.40   & 65.82 & 67.11 & 69.09 & 60.11 \\  
		{RseNet-18 (ours, w/o pre-train)} & 77.18 & 72.75  & 57.00   & 62.65   & 91.00 & 89.60 & 88.40 & 78.85 \\  
        {RseNet-50 (ours, w pre-train)} & \underline{86.26} & \underline{89.45}  & \underline{66.60}   & \underline{67.95}   & \underline{97.85} & \underline{98.15} & \underline{96.60} & \underline{87.20} \\ 
        CoPINet            & \bf{91.42} & \bf{95.05}  & \bf{77.45}   & \bf{78.85}   & \bf{99.10} & \bf{99.65} & \bf{98.50} & \bf{91.35} \\ \hline  
		Human              & 84.41 & 95.45  & 81.82   & 79.55   & 86.36 & 81.81 & 86.36 & 81.81 \\ \bottomrule
	\end{tabular}
	\label{tbl_diffnet}
	\vspace{-1em}
\end{table*}

\subsection{Implementation Details}
We follow the experimental setup in~\cite{CVPR2019_Zhang}. The RAVEN dataset is split into three parts, 6 folds for training, 2 for validation and the rest 2 for testing. We train our models on the training set, tune the model parameters on the validation set and report accuracy on the test set.

We use ResNet as the backbone in all experiments. Besides, we freeze the parameters of all batch normalization~\cite{ICML2015_Ioffe} layers during training. To  reduce the effect of over-fitting, we add an extra dropout layer~\cite{JMLR2014_Srivastava} with a default value (set to 0.5) before the fully connected layer in both supervised and unsupervised models. 

For model training, we use a mini-batch of 32 with Adam optimizer \cite{ICLR2015_Adam} to learn network parameters. The learning rate is set to 0.0002 and is reduced to its half every 10 iterations. In addition, as the standard image resolution in ResNet~\cite{CVPR2016_He}, all images in the problem matrix and answer set are resized to a fixed size of $224 \times 224$ and are normalized to a range of $[0, 1]$. To make a fair comparison to the previous work~\cite{CVPR2019_Zhang}, we do not use any data augmentation method to increase the diversity of training samples.

\subsection{Supervised Learning Approach} 

\subsubsection{Comparison with Baselines} 
To demonstrate the effectiveness of our method, we report all available results on the RAVEN dataset~\cite{CVPR2019_Zhang} for comparison, which include LSTM~\cite{NIPs2015_Shi}, CNN~\cite{arxiv2017_Hoshen}, ResNet~\cite{CVPR2016_He}, WReN~\cite{ICML2018_Santoro} and DRT~\cite{CVPR2019_Zhang}. Detailed implementations of these baselines are described in~\cite{CVPR2019_Zhang}. In addition, we compare with the most recent algorithm CoPINet~\cite{NeurIPS2019_Zhang}.

\tab \ref{tbl_diffnet} shows the testing accuracy of each model on the RAVEN dataset.\footnote{To clearly describe the difference among multiple ResNet backbone based models, we use ResNet-18+MLP to represent ResNet and use ResNet-18+MLP+DRT to denote the best performing model DRT in~\cite{CVPR2019_Zhang}.} We can observe that the performance for RPM heavily relies on the network architecture. LSTM and WReN perform poorly on this dataset. They are slightly better than random guessing (12.5\%). CNN model obtains an accuracy of 36.97\%, which is still not good enough. Based on the ResNet backbone, ResNet-18+MLP+DRT obtains an increase of about 6\% when compared to ResNet-18+MLP. In contrast, our ResNet-18 outperforms ResNet-18+MLP+DRT by a large margin of 17.62\%. The reason for such results is adding of additional MLP modules to ResNet architecture may reduce the testing accuracy. 

In \tab \ref{tbl_diffnet}, only two models surpass the human-level performance, \ie CoPINet (91.42\%) and our ResNet-50 (86.26\%). The average accuracy of our model is about 5\% less. Notice that there is not any additional network design in our model. We only use a simple ResNet-50 architecture with ImageNet pre-training, verifying the effectiveness of model pre-training and deep layers in supervised learning. We hope these two strategies can help researchers in their future works on RPM problems. Detailed comparison with human-level performance will be discussed in Section \ref{sec_machInt}.

\subsubsection{Generalization Test}
\label{subsec_gen}

To measure how well a supervised learning model trained on one figure configuration and tested on other similar configurations, we further test the generalization ability of our approach on the RAVEN dataset. Similar to the experimental setup in DRT~\cite{CVPR2019_Zhang}, we evaluate the supervised model (best performing model with ResNet-50 backbone) with three kinds of configurations:
\begin{itemize}
	\item Train on \emph{Center} and test on \emph{L-R}, \emph{U-D} and \emph{O-IC}. This test requires the model to learn the rules on a single component, but generalize to multiple similar components, see \tab \ref{tbl_center}.
	
	\item Train on \emph{L-R} and test on \emph{U-D}, and vice versa. This test can measure the transposition performance of a model, since \emph{L-R} and \emph{U-D} can be transposed to each other, see \tab \ref{tbl_lu}.
	
	\item Train on \emph{2*2Grid} and test on \emph{3*3Grid}, and vice-versa. This test evaluates the generalization when the number of objects changes, see \tab \ref{tbl_g23}.
\end{itemize}

\begin{table}[]
	\centering
	\caption{Generalization test. The model is trained on \emph{Center} and tested on three other figure configurations.}
	\begin{tabular}{ccccc} \toprule	
	Method     & \emph{Center} & \emph{L-R} & \emph{U-D} & \emph{O-IC} \\ \midrule
	DRT        & 51.87  & 40.03      & 35.46    & 38.84   \\ 
	Ours       & \bf{60.80}  & \bf{43.65}  & \bf{41.40}  & \bf{43.65}   \\   	\bottomrule
	\end{tabular}
	\label{tbl_center}
\end{table}
\begin{table}[]
	\centering
	\caption{Generalization test. The mode is trained on \emph{L-R} and tested on \emph{U-D}, and vice versa.}
	\begin{tabular}{ccccc}
		\toprule
		\multirow{2}{*}{Config} & \multicolumn{2}{c}{DRT} & \multicolumn{2}{c}{Ours} \\ \cmidrule(lr){2-3} \cmidrule(lr){4-5} 
		& \emph{L-R}          & \emph{U-D}         & \emph{L-R}          & \emph{U-D}          \\ \midrule
		\emph{L-R} &	41.07       & 38.10       & \bf{60.95}       & \bf{57.15}       \\ 
		\emph{U-D} &    39.48       & 43.60       & \bf{59.20}       & \bf{63.75}       \\ \bottomrule
	\end{tabular}
	\label{tbl_lu}
\end{table}

\begin{table}[]
	\centering
	\caption{Generalization test. The mode is trained on \emph{2*2Grid} and tested on \emph{3*3Grid}, and vice versa.}
	\begin{tabular}{ccccc}
		\toprule
		\multirow{2}{*}{Config} & \multicolumn{2}{c}{DRT} & \multicolumn{2}{c}{Ours} \\ \cmidrule(lr){2-3} \cmidrule(lr){4-5} 
	     	& \emph{2*2Grid}          & \emph{3*3Grid}         & \emph{2*2Grid}          & \emph{3*3Grid}          \\ \midrule
		\emph{2*2Grid}	&	\bf{40.93}       & \bf{38.69}       & 35.90       & 35.55       \\ 
		\emph{3*3Grid}	&	\bf{39.14}       & \bf{43.72}       & 34.70       & 40.05       \\ \bottomrule
	\end{tabular}
	\label{tbl_g23}
\end{table}

Tables \ref{tbl_center}, \ref{tbl_lu}, and \ref{tbl_g23} are the results of generalization tests of our ResNet-50 (denoted as Ours) and ResNet-18+MLP+DRT (denoted as DRT).\footnote{The generalization tests are not conducted in CoPINet~\cite{NeurIPS2019_Zhang}.} Compared to the results reported in \tab \ref{tbl_diffnet}, we can observe that training on a single figure configuration cannot achieve better performance than training on multiple figure configurations. Increasing the diversity of figure configurations could improve the model generalization and performance, although the rules in those figure configurations are different. Such results are observed in both DRT~\cite{CVPR2019_Zhang} and our method.
	
\tab \ref{tbl_center} indicates that it is difficult to answer complex RPM questions by learning a model on simple cases. The three other configurations (\ie \emph{L-R}, \emph{U-D} and \emph{O-IC}) consist of multiple \emph{Center} components, and the accuracy on these configurations heavily drops, up to 16.41\% in DRT and 19.40\% in ours. Such results are reasonable as we should not expect a student from a junior level (\eg primary school) to pass the examination of a senior level (\eg high school).
	
In \tab \ref{tbl_lu}, the two figure configurations can be considered as transpose of each other. Similarly, in \tab \ref{tbl_g23}, \emph{3*3Grid} is an simple extension of \emph{2*2Grid} patterns, except the number of objects. When transferring the model to similar patterns, the performance of both our method and DRT does not change much. An interesting observation is: in Tables \ref{tbl_center}, \ref{tbl_lu}, our model outperforms DRT on all configurations, but it performs worse than DRT on \emph{2*2Grid} and \emph{3*3Grid}, although the overall performance of our model is much better than DRT, \ie 86.26 vs 59.56\% (see \tab \ref{tbl_diffnet}). Such an observation indicates that model generalization of supervised learning is very challenging.

\begin{table*}[!t]
	\centering
	\caption{Testing accuracy of our unsupervised method on the test set of the RAVEN dataset.}
	\begin{tabular}{lcccccccc} \toprule
	Method             & \emph{Avg}   & \emph{Center} & \emph{2*2Grid}  & \emph{3*3Grid} 
	 & \emph{L-R}   & \emph{U-D}   & \emph{O-IC}  & \emph{O-IG}  \\  \midrule
	Random         & 12.50 & 12.50  & 12.50  & 12.50   & 12.50  & 12.50  & 12.50  & 12.50 \\ 
	w/o MCPT         & 5.18  & 7.60  & 7.00  & 7.00   & 2.45  & 3.20  & 3.20  & 5.30  \\
	MCPT (one-hot) & 20.55  & 26.60  & 20.80  & 22.15   & 18.25  & 17.30  & 21.65  & 17.10  \\
	%MCPT (two-hot)   & {\bf 27.86}  & {\bf 35.30}  & {\bf 24.95} &  {\bf 26.50}  &  {\bf 28.55} & {\bf 26.55} & {\bf 32.10} & {\bf 21.05} 	\\  \bottomrule
	MCPT (two-hot)    & {\bf 28.50}  & {\bf 35.90}  & {\bf 25.95}   & {\bf 27.15}  & {\bf 29.30}   & {\bf 27.40}   & {\bf 33.10}   & {\bf 20.70} \\ \bottomrule   
	\end{tabular} 
	\label{tbl_unsup}
\end{table*}

\subsection{Unsupervised Learning Approach}
After testing several ResNet variants, we chose the ResNet-18 backbone in our unsupervised model for its good performance.\footnote{For unsupervised learning approaches, the target to be learned is unknown. Although we use a pseudo target in this work, it is a weak prior only and is far from the true target. Without an appropriate supervision, a deep neural network with many layers would easily be trapped in a local minimum, because of too many parameters.} Since there is no existing algorithm that solves RPM in an unsupervised manner, we design two types of baselines to verify the effectiveness of our MCPT approach. 
\begin{itemize}
	\item Random guess. Without any background knowledge on RPM, random guess is a natural baseline for answer prediction. As there are 8 candidate choices in the answer set, the average accuracy of random guess on each figure configuration is 12.5\%. 
	
	\item To verify the effectiveness of pseudo target (two-hot vector), we report the accuracy of predefined model without pseudo target, and model training with one-hot (the first element is set to 1 only) pseudo target. In our experiments, the weights of ResNet-18 backbone is initialized by the ImageNet pre-training, and the fully connected layer is randomly initialized. All parameters in these models are kept the same.
\end{itemize}

\tab \ref{tbl_unsup} shows the testing accuracy of our unsupervised methods on the test set of the RAVEN dataset. From \tab \ref{tbl_unsup}, it can be seen that the average accuracy is only 5.18\% without pseudo target based learning, which is even worse than the result of random guessing. With the pseudo target of a one-hot vector, the testing accuracy is effectively improved by a large margin of 15.37\%. Furthermore, based on the proposed pseudo target of a two-hot vector, the performance is further improved by about 7.95\%. Based to above results, the effectiveness of MCPT is promising.

In addition, when compared to the random guess strategy, our MCPT (two-hot) achieves an average accuracy of 28.50\%, which doubles the performance. Specifically, our unsupervised method outperforms random guess on all 7 figure configurations, and achieves a testing accuracy of 35.90\% on the \emph{Center} and 33.10\% on \emph{O-IC} configurations. Even when compared to some supervised methods, \eg, LSTM (13.07\%) and WReN (14.69\%), our unsupervised method performs better, see \tab \ref{tbl_diffnet}. However, compared to the human-level performance (84.41\%), the unsupervised approach still has a long way to go.

\section{Machine Intelligence}
\label{sec_machInt}
In the past five years, machines have defeated humans in many tasks, such as image classification~\cite{ICCV2015_He} and face verification~\cite{AAAI2015_Lu}. Unlike those tasks, RPM is highly correlated with human intelligence. One natural question arises: have machines attained reasoning ability comparable to that of humans even in such limited settings? To answer this question, we first summarize our observations based on the extensive results in Section \ref{sec_exp}, and then we discuss  machine intelligence by comparing those observations with the human-level performance.

{\bf Supervised learning.} \tab \ref{tbl_diffnet} shows that CoPINet~\cite{NeurIPS2019_Zhang} and our approach outperform the human-level performance on the RAVEN dataset. From \tab \ref{tbl_diffnet}, we observe consistent results between CoPINet and our method, although they are different models. First, these two methods surpass humans on four figure configurations, \ie \emph{L-R}, \emph{U-D}, \emph{O-IC}, and \emph{O-IG}. Specifically, these two approaches outperform humans by a significant margin of more than 10\% on the \emph{L-R} and \emph{O-IC} figure configuration. Especially, on the figure configuration \emph{U-D}, more than 16\% increase is obtained. Second, for the rest of the configurations, \ie \emph{Center}, \emph{2*2Grid}, and \emph{3*3Grid}, humans perform better than these two methods. \tab \ref{tbl_diffnet} also shows the model bias on some figure configurations.

In addition, as shown in the results of generalization test reported in Section \ref{subsec_gen}, the performance of supervised learning heavily relies on the training data. When the model is trained on one figure configuration and tested on others, it is difficult to generalize to new problems, Especially, when the model is trained on simple cases but tested on complicated ones, this phenomenon becomes more obvious, as the results reported in \tab \ref{tbl_center}. In contrast, humans are good at solving complex problems after training on basic cases.

Furthermore, although CoPINet and our approach obtain better average accuracy than humans on the RAVEN dataset, the logical rules hidden in each RPM problem are still unknown for machines. As illustrated in \fig \ref{fig_rpm}, given an RPM problem, humans can extract these hidden rules to explain how such a result is obtained, while machines only predict the final answer.

{\bf Unsupervised learning.}
Compared to the supervised learning strategy, unsupervised approaches aim at automatically solving RPM without any supervision. In \tab \ref{tbl_unsup}, the proposed MCPT approach obtains a testing accuracy of 28.50\% without any labeled data. Compared to the human-level performance, unsupervised approaches still have a big gap to bridge. However, for some configurations, \ie \emph{Center} and \emph{O-IC}, the testing accuracy of MCPT is more than 33\%. In addition, double accuracy of the random guess strategy also indicates the potential of unsupervised learning. 

{\bf Discussions.} Based on above observations, we speculate on two interesting questions as follows:
\begin{itemize}	
	%\item {\bf Is good reasoning accuracy equal to intelligence?} No, unless it is also explainable. Intelligent systems are not only required to provide an answer to a question, but also explain how such an answer is obtained. Although the supervised models slightly outperform human-level performance, they only predict the answer. Logical rules (as in \fig \ref{fig_rpm}) hidden in those RPM problems are still unknown to machines.
			
	\item {\bf Is unsupervised learning method necessary?} Yes. Based on results presented in Tables 2-5, although supervised learning methods can obtain good results, their performance heavily relies on the training data. It is also hard to generalize them to new problems. For an intelligent system, it is expected to rapidly generalize to new tasks and situations, and learn new skills to solve new problems. Therefore, unsupervised learning method needs further exploration. 
	
	\item {\bf How to learn logical rules in unsupervised learning?} As illustrated in \fig \ref{fig_rpm}, the logical rules hidden in different RPM problems are manifested as different visual structures. If we can build a grammar tree to represent those visual structures, recovering logical rules is possible. 
	Based on such a strategy, unsupervised representation learning on visual structures would be a potential solution. Then, the model of unsupervised representation learning is explainable.

\end{itemize}

\section{Conclusion}
To study machine intelligence, we attempt to solve RPM in both supervised and unsupervised manners. To reduce over-fitting in supervised learning, we suggest the use of a neural network with deep layers and pre-training on large scale datasets. Extensive experiments on the RAVEN dataset verify the 
effectiveness of 
these two strategies. We further investigate the unsupervised learning strategy that attempts to be more human-like. Without a rule-based feature extraction method in advance, we propose a novel method MCPT that converts the unsupervised strategy into a supervised one. Experiments show that our unsupervised method doubles the performance of random guess, which demonstrates its potential. Finally, we discuss the need for more works on unsupervised and explainable strategies.

\clearpage

{\small
\bibliographystyle{ieee_fullname}
\bibliography{egbib}
}

\end{document}